\documentclass[conference,letterpaper]{IEEEtran}
\IEEEoverridecommandlockouts

\usepackage[T1]{fontenc}
\usepackage{mathptmx}
\usepackage{amsmath,amssymb,amsfonts}
\usepackage{algorithmic}
\usepackage{algorithm}
\usepackage{graphicx}
\usepackage{textcomp}
\usepackage{booktabs}
\usepackage{multirow}
\usepackage{tikz}
\usepackage{pgfplots}
\pgfplotsset{compat=1.18}
\usetikzlibrary{shapes.geometric, arrows.meta, positioning, fit, backgrounds, calc}
\usepackage{xcolor}
\usepackage{url}
\usepackage{cite}
\usepackage{balance}

\definecolor{sysblue}{RGB}{41,98,255}
\definecolor{sysgreen}{RGB}{0,150,80}
\definecolor{sysorange}{RGB}{230,126,34}
\definecolor{sysred}{RGB}{192,57,43}
\definecolor{syspurple}{RGB}{142,68,173}
\definecolor{lightgray}{RGB}{245,245,245}

\title{MetaCogAgent: A Metacognitive Multi-Agent LLM Framework with Self-Aware Task Delegation}

\author{
\IEEEauthorblockN{Chenyu Wang\textsuperscript{1} and Yang Shu\textsuperscript{2,*}}
\IEEEauthorblockA{\textsuperscript{1}School of Computer and Artificial Intelligence, Zhengzhou University, Zhengzhou, China\\
\textsuperscript{2}Zhejiang University, Hangzhou, China\\
Email: 18624902337@163.com, shuyang@zju.edu.cn}
\thanks{*Corresponding author: Yang Shu (shuyang@zju.edu.cn)}
}

\begin{document}

\maketitle

\begin{abstract}
Multi-agent large language model (LLM) systems have shown promise for solving complex tasks through agent collaboration. However, existing frameworks assign tasks based on predefined roles without considering whether an agent can accurately assess its own competence boundaries, leading to overconfident execution on tasks beyond its expertise. Inspired by metacognition theory from cognitive science, we propose \textbf{MetaCogAgent}, a multi-agent LLM framework where each agent is equipped with a \emph{Metacognitive Self-Assessment Unit} that evaluates task-capability alignment before execution. The framework introduces three contributions: (1)~a \textbf{self-assessment mechanism} that estimates per-task confidence by combining verbalized uncertainty with historical capability profiles; (2)~an \textbf{adaptive delegation protocol} that routes low-confidence tasks to better-suited agents through cross-agent evaluation; and (3)~a \textbf{capability boundary learning} module that iteratively refines each agent's competence model via cybernetic feedback. Experiments on our constructed \emph{MetaCog-Eval} benchmark (700 tasks across 5 cognitive dimensions) demonstrate that MetaCogAgent achieves 82.4\% task accuracy---8.7\% above the best routing baseline---while using 5\% fewer API calls than AutoGen and 34\% fewer than ensemble voting. Ablation studies confirm that each metacognitive component contributes to overall system performance.
\end{abstract}


\section{Introduction}

Large language models (LLMs) have demonstrated remarkable capabilities across diverse tasks~\cite{brown2020language, achiam2023gpt4}. To tackle complex problems requiring heterogeneous skills, recent work has explored \emph{multi-agent LLM systems}, where multiple specialized agents collaborate through structured communication~\cite{wu2023autogen, hong2023metagpt, li2023camel, park2023generative}. These systems decompose complex tasks into subtasks and assign them to agents with designated roles (e.g., ``coder,'' ``researcher,'' ``critic'').

However, current multi-agent frameworks suffer from a fundamental limitation: \textbf{agents lack self-awareness of their own competence boundaries}. When a reasoning-specialized agent receives a coding subtask, it does not recognize the mismatch---it simply attempts execution with full confidence, often producing plausible but incorrect results. This ``metacognitive blindness'' leads to cascading errors in multi-agent pipelines, as downstream agents build upon flawed outputs without detecting upstream failures.

The concept of \emph{metacognition}---``thinking about one's own thinking''---has been extensively studied in cognitive science~\cite{flavell1979metacognition, toppino2009metacognitive}. Humans routinely assess their own knowledge and abilities, recognizing when a task exceeds their competence and seeking help from others. This metacognitive capacity is precisely what current multi-agent LLM systems lack.

We propose \textbf{MetaCogAgent}, a metacognitive multi-agent framework that endows each LLM agent with three capabilities inspired by human metacognition:

\begin{enumerate}
    \item \textbf{Self-Assessment}: Before executing a task, each agent evaluates its confidence through verbalized uncertainty estimation combined with a learned capability profile, producing a calibrated competence score.
    \item \textbf{Adaptive Delegation}: When self-assessed confidence falls below a threshold, the agent initiates a delegation protocol---broadcasting the task to peer agents for cross-evaluation and routing it to the most competent candidate.
    \item \textbf{Capability Boundary Learning}: After task completion, performance feedback updates each agent's capability profile through a cybernetic loop, progressively refining competence boundaries over time.
\end{enumerate}

To evaluate metacognitive capabilities---which existing benchmarks do not measure---we construct \emph{MetaCog-Eval}, a benchmark of 700 tasks spanning five cognitive dimensions (reasoning, retrieval, coding, mathematics, commonsense) with annotated difficulty levels and optimal agent assignments.

Experiments show that MetaCogAgent achieves 82.4\% accuracy on MetaCog-Eval, outperforming the strongest routing baseline by 8.7\% while using 5\% fewer API calls than AutoGen and 34\% fewer than ensemble methods. The self-assessment module achieves an Expected Calibration Error (ECE) of 0.087, indicating that agents can reliably estimate their own competence.

\section{Related Work}

\textbf{Multi-Agent LLM Systems.}
Recent frameworks enable multiple LLM agents to collaborate on complex tasks. AutoGen~\cite{wu2023autogen} provides a conversation-based framework for customizable agent interactions. MetaGPT~\cite{hong2023metagpt} assigns software engineering roles to agents following a standardized operating procedure. CAMEL~\cite{li2023camel} explores communicative agents through role-playing. AgentVerse~\cite{chen2024agentverse} studies emergent collaborative behaviors. Multi-agent debate~\cite{du2023improving} and exchange-of-thought~\cite{yin2023exchange} improve reasoning through inter-agent discussion. However, \emph{none of these works} address whether agents can assess their own capability boundaries or adaptively delegate tasks they cannot handle.

\textbf{LLM Confidence and Calibration.}
Kadavath et al.~\cite{kadavath2022language} demonstrate that LLMs exhibit some ability to ``know what they know,'' though confidence is often poorly calibrated. Xiong et al.~\cite{xiong2024llms} evaluate confidence elicitation strategies including verbalized confidence, consistency-based methods, and hybrid approaches. Guo et al.~\cite{guo2017calibration} establish the Expected Calibration Error (ECE) metric for neural network confidence assessment. These works study confidence at the single-model level; we extend this to multi-agent settings where confidence informs delegation decisions.

\textbf{Metacognition in AI.}
While metacognition is a foundational concept in cognitive science~\cite{flavell1979metacognition}, its application to AI systems remains nascent. Reflexion~\cite{shinn2023reflexion} enables agents to reflect on failures for future improvement, representing a form of retrospective metacognition. Tree of Thoughts~\cite{yao2023tree} enables deliberate exploration of reasoning paths, incorporating an implicit form of self-evaluation. Our work differs by implementing \emph{prospective metacognition}---assessing competence \emph{before} task execution to prevent failures rather than learning from them after the fact.

\textbf{Cybernetic Systems.}
The Systems, Man, and Cybernetics (SMC) community has long studied feedback-driven self-regulating systems~\cite{wiener1948cybernetics}. Our capability boundary learning module directly instantiates the cybernetic feedback loop: agent performance generates error signals that update the internal model (capability profile), which in turn improves future control decisions (delegation). This positions MetaCogAgent as a cybernetic multi-agent system where self-awareness emerges from feedback-driven adaptation.

\section{MetaCogAgent Framework}

\subsection{System Overview}

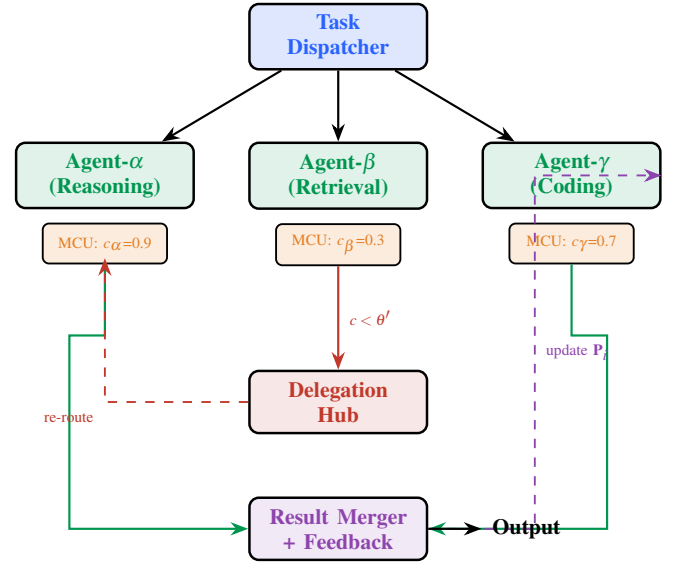
\begin{figure}[t]
\centering
\resizebox{\columnwidth}{!}{
\begin{tikzpicture}[
    node distance=0.6cm and 0.8cm,
    box/.style={rectangle, draw, rounded corners=3pt, minimum width=2.0cm, minimum height=0.7cm, font=\scriptsize\bfseries, align=center, line width=0.8pt},
    smallbox/.style={rectangle, draw, rounded corners=2pt, minimum width=1.4cm, minimum height=0.45cm, font=\tiny, align=center, line width=0.6pt},
    arrow/.style={-{Stealth[length=2mm]}, line width=0.7pt},
    dasharrow/.style={-{Stealth[length=2mm]}, dashed, line width=0.6pt, color=gray},
]

\node[box, fill=sysblue!15, text=sysblue] (dispatch) {Task\\Dispatcher};

\node[box, fill=sysgreen!12, text=sysgreen, below left=0.8cm and 0.6cm of dispatch] (agentA) {Agent-$\alpha$\\(Reasoning)};
\node[box, fill=sysgreen!12, text=sysgreen, below=0.8cm of dispatch] (agentB) {Agent-$\beta$\\(Retrieval)};
\node[box, fill=sysgreen!12, text=sysgreen, below right=0.8cm and 0.6cm of dispatch] (agentC) {Agent-$\gamma$\\(Coding)};

\node[smallbox, fill=sysorange!15, text=sysorange, below=0.15cm of agentA] (mcuA) {MCU: $c_\alpha$=0.9};
\node[smallbox, fill=sysorange!15, text=sysorange, below=0.15cm of agentB] (mcuB) {MCU: $c_\beta$=0.3};
\node[smallbox, fill=sysorange!15, text=sysorange, below=0.15cm of agentC] (mcuC) {MCU: $c_\gamma$=0.7};

\node[box, fill=sysred!12, text=sysred, below=1.8cm of agentB] (delegation) {Delegation\\Hub};

\node[box, fill=syspurple!12, text=syspurple, below=0.7cm of delegation] (merger) {Result Merger\\+ Feedback};

\draw[arrow] (dispatch) -- (agentA);
\draw[arrow] (dispatch) -- (agentB);
\draw[arrow] (dispatch) -- (agentC);

\draw[arrow, color=sysred] (mcuB.south) -- node[right, font=\tiny, color=sysred]{$c < \theta'$} (delegation);

\draw[arrow, color=sysgreen] (mcuA.south) |- +(-0.4,-0.8) |- (merger.west);
\draw[arrow, color=sysgreen] (mcuC.south) |- +(0.4,-0.8) |- (merger.east);

\draw[dasharrow, color=sysred] (delegation.west) -| node[below left, font=\tiny]{re-route} ($(agentA.south)+(0,-0.55)$);

\draw[dasharrow, color=syspurple] (merger.east) -- +(1.2,0) |- node[right, font=\tiny, near start]{update $\mathbf{P}_i$} (agentC.east);

\node[right=0.6cm of merger, font=\scriptsize\bfseries] (output) {Output};
\draw[arrow] (merger) -- (output);

\end{tikzpicture}
}
\caption{MetaCogAgent architecture. Each agent's Metacognitive Unit (MCU) produces confidence scores. Low-confidence tasks ($c < \theta'$) are routed to the Delegation Hub for reassignment. The Result Merger provides feedback to update capability profiles $\mathbf{P}_i$.}
\label{fig:architecture}
\end{figure}

MetaCogAgent consists of $N$ specialized LLM agents $\{A_1, A_2, \ldots, A_N\}$, a \emph{Task Dispatcher}, a \emph{Delegation Hub}, and a \emph{Result Merger}. Each agent $A_i$ is equipped with a \emph{Metacognitive Unit} (MCU) comprising three components: a Self-Assessment module, a Capability Profile $\mathbf{P}_i$, and a feedback interface. The system architecture is illustrated in Fig.~\ref{fig:architecture}.

\subsection{Metacognitive Self-Assessment}

When agent $A_i$ receives subtask $t_k$, its MCU estimates a confidence score $c_i(t_k) \in [0, 1]$ before execution. The score integrates two signals:

\textbf{Verbalized Confidence ($c^v$).} The agent is prompted to assess its competence for the given task:

\vspace{-2mm}
\begin{quote}
\emph{``Before solving this task, rate your confidence (0--100) based on: (1) whether this matches your expertise, (2) how certain you are about the approach, (3) whether you have sufficient knowledge.''}
\end{quote}
\vspace{-2mm}

The agent responds with a structured JSON containing a numeric score, which is parsed and normalized to $c^v_i(t_k) \in [0,1]$.

\textbf{Profile-Based Confidence ($c^p$).} Each agent maintains a capability profile $\mathbf{P}_i = [p_{i,1}, p_{i,2}, \ldots, p_{i,D}]$ where $p_{i,d}$ represents the agent's historical success rate on dimension $d$ (e.g., reasoning, coding). The dimension label $d_k$ is extracted from the task description via a lightweight LLM classifier prior to dispatch. Given $d_k$:
\begin{equation}
    c^p_i(t_k) = p_{i,d_k}
\end{equation}

The final confidence score combines both signals:
\begin{equation}
    c_i(t_k) = \lambda \cdot c^v_i(t_k) + (1 - \lambda) \cdot c^p_i(t_k)
    \label{eq:confidence}
\end{equation}
where $\lambda \in [0,1]$ balances verbalized and profile-based estimates. Higher values favor verbalized confidence (useful when profiles are uninformed), while lower values emphasize historical performance.

\textbf{Metacognitive Conflict Detection.} Beyond the composite score $c_i(t_k)$, the \emph{disagreement} between the two signals, $\delta_i(t_k) = |c^v_i(t_k) - c^p_i(t_k)|$, provides a second-order metacognitive indicator---capturing epistemic uncertainty about the self-assessment itself. A large $\delta$ signals that the agent's real-time introspection diverges from its historical track record: the agent may encounter a task type not yet reflected in its profile (high $c^v$, low $c^p$), or it may be verbally uncertain despite historical competence (high $c^p$, low $c^v$), indicating potential distribution shift. When $\delta_i$ exceeds a conflict threshold $\theta_\delta$ (set to 0.3), the delegation threshold is tightened to $\theta' = \theta + \gamma \cdot \delta_i$ with dampening factor $\gamma = 0.2$, making the system more conservative under metacognitive uncertainty. This mechanism ensures that high-composite-confidence assessments are treated with appropriate caution when the underlying signals conflict---a form of \emph{second-order self-doubt} inspired by metacognitive monitoring research~\cite{toppino2009metacognitive}.

\subsection{Adaptive Delegation Protocol}

When $c_i(t_k) < \theta'$ (the effective delegation threshold, adjusted by conflict detection per Section~III-B), agent $A_i$ triggers the delegation protocol:

\begin{enumerate}
    \item \textbf{Broadcast}: Task $t_k$ is sent to all other agents $A_j$ ($j \neq i$) for confidence evaluation.
    \item \textbf{Cross-Evaluation}: Each candidate agent $A_j$ computes $c_j(t_k)$ using its own MCU.
    \item \textbf{Assignment}: The task is assigned to $A^* = \arg\max_j c_j(t_k)$.
    \item \textbf{Collaborative Fallback}: If $\max_j c_j(t_k) < \theta$ for all agents, the system enters \emph{Collaborative Mode}: all agents independently solve $t_k$, and outputs are aggregated via weighted voting:
\end{enumerate}
\begin{equation}
    \hat{y}_k = \arg\max_y \sum_{i=1}^{N} c_i(t_k) \cdot \mathbb{1}[y_i = y]
    \label{eq:voting}
\end{equation}

The delegation protocol's computational overhead is bounded: it requires at most $N-1$ additional confidence evaluations per delegated task, each consisting of a single short LLM inference (no full task execution). Algorithm~\ref{alg:metacog} summarizes the full pipeline.

\begin{algorithm}[t]
\caption{MetaCogAgent Task Processing}
\label{alg:metacog}
\scriptsize
\begin{algorithmic}[1]
\REQUIRE Task $t_k$, agents $\{A_1, \ldots, A_N\}$, threshold $\theta$, $\theta_\delta$, $\gamma$
\STATE Dispatcher assigns $t_k$ to agent $A_i$ via round-robin
\STATE $c_i \leftarrow \lambda \cdot c^v_i(t_k) + (1-\lambda) \cdot c^p_i(t_k)$ \hfill $\triangleright$ Self-assess
\STATE $\delta_i \leftarrow |c^v_i - c^p_i|$; ~$\theta' \leftarrow \theta + \gamma \cdot \delta_i \cdot \mathbb{1}[\delta_i > \theta_\delta]$ \hfill $\triangleright$ Conflict-aware
\IF{$c_i \geq \theta'$}
    \STATE $y_k \leftarrow A_i.\text{execute}(t_k)$ \hfill $\triangleright$ Direct execution
\ELSE
    \FOR{each $A_j$ where $j \neq i$}
        \STATE $c_j \leftarrow \lambda \cdot c^v_j(t_k) + (1-\lambda) \cdot c^p_j(t_k)$
    \ENDFOR
    \IF{$\max_j c_j \geq \theta$}
        \STATE $A^* \leftarrow \arg\max_j c_j(t_k)$
        \STATE $y_k \leftarrow A^*.\text{execute}(t_k)$ \hfill $\triangleright$ Delegate
    \ELSE
        \STATE $y_k \leftarrow \text{WeightedVote}(\{A_j\}, \{c_j\}, t_k)$ \hfill $\triangleright$ Collaborate
    \ENDIF
\ENDIF
\STATE $r_k \leftarrow \text{evaluate}(y_k)$
\STATE Update $p_{i,d_k}$ via Eq.~\eqref{eq:update} \hfill $\triangleright$ Boundary learning
\RETURN $y_k$
\end{algorithmic}
\end{algorithm}

\subsection{Capability Boundary Learning}

After the Result Merger evaluates output quality for task $t_k$ (via exact-match against ground truth for objective tasks, or GPT-4 judge scoring for open-ended tasks), it updates the executing agent's capability profile:
\begin{equation}
    p_{i,d_k}^{(t+1)} = p_{i,d_k}^{(t)} + \alpha \cdot \left( r_k - p_{i,d_k}^{(t)} \right)
    \label{eq:update}
\end{equation}
where $r_k \in \{0, 1\}$ is the correctness indicator and $\alpha = 0.1$ is the learning rate. This exponential moving average formulation mirrors the update rule in classical adaptive control systems~\cite{wiener1948cybernetics}, where the capability profile serves as the system's internal model, performance feedback acts as the error signal, and profile updates close the cybernetic loop.

Over successive tasks, each agent's profile $\mathbf{P}_i$ progressively approximates its operational competence boundaries, enabling increasingly informed delegation decisions.

\textbf{Bayesian Interpretation.} The EMA update has an intuitive connection to conjugate Bayesian estimation: modeling task success as Bernoulli with a Beta prior, the EMA approximates the posterior mean with an effective memory horizon of $n_{\text{eff}} \approx 1/\alpha$ tasks. With $\alpha = 0.1$, each observation carries the influence of $\sim$10 recent tasks, automatically discounting obsolete data. The sensitivity analysis (Table~\ref{tab:sensitivity}) empirically confirms that $\alpha = 0.1$ balances adaptivity and robustness.

\section{MetaCog-Eval Benchmark}

Existing multi-agent evaluation benchmarks measure task completion accuracy but do not assess agents' metacognitive capabilities---specifically, whether agents can accurately judge \emph{what they can and cannot do}. We construct MetaCog-Eval to fill this gap.

\subsection{Benchmark Design}

MetaCog-Eval consists of 700 tasks across five cognitive dimensions: \emph{Logical Reasoning} (LR), \emph{Knowledge Retrieval} (KR), \emph{Code Generation} (CG), \emph{Mathematical Computation} (MC), and \emph{Commonsense Inference} (CI). Each dimension contains tasks at three difficulty levels (Easy/Medium/Hard, 40 tasks each), plus 100 \emph{cross-domain} tasks requiring competencies from multiple dimensions simultaneously.

Tasks are generated using GPT-4 with dimension-specific prompts, then filtered and validated by two human annotators. Each task is labeled with: (a) ground-truth answer, (b) required capability dimension(s), (c) difficulty level, and (d) optimal agent assignment. Inter-annotator agreement reaches $\kappa = 0.81$ (substantial agreement). Table~\ref{tab:dataset} summarizes the benchmark statistics.

To illustrate the range of metacognitive challenges, consider two examples. An Easy-LR task might ask: ``\textit{If all roses are flowers and some flowers fade quickly, can we conclude all roses fade quickly?}'' where logical structure is transparent. A Hard cross-domain task might be: ``\textit{Write a simulation of the Monty Hall problem in Python, run 10{,}000 trials, and explain why the empirical probability converges to 2/3}''---requiring coding, mathematical reasoning, and probabilistic intuition simultaneously. The latter demands accurate self-assessment: an agent strong in coding but weak in probability theory should recognize the mismatch and delegate.

\begin{table}[t]
\centering
\caption{MetaCog-Eval benchmark statistics.}
\label{tab:dataset}
\scriptsize
\begin{tabular}{lccccc}
\toprule
\textbf{Dimension} & \textbf{Easy} & \textbf{Med.} & \textbf{Hard} & \textbf{Cross} & \textbf{Total} \\
\midrule
Logical Reasoning & 42 & 38 & 40 & --- & 120 \\
Knowledge Retrieval & 38 & 42 & 40 & --- & 120 \\
Code Generation & 40 & 38 & 42 & --- & 120 \\
Math Computation & 41 & 40 & 39 & --- & 120 \\
Commonsense Inf. & 39 & 42 & 39 & --- & 120 \\
Cross-domain & --- & --- & --- & 100 & 100 \\
\midrule
\textbf{Total} & 200 & 200 & 200 & 100 & \textbf{700} \\
\bottomrule
\end{tabular}
\end{table}

\section{Experiments}

\subsection{Setup}

We instantiate MetaCogAgent with $N=3$ specialized agents: Agent-$\alpha$ (reasoning-focused), Agent-$\beta$ (retrieval-focused), and Agent-$\gamma$ (coding-focused). Each agent is a prompted GPT-4 instance with distinct system instructions. The dispatcher assigns tasks via round-robin (without knowledge of task content), so delegation gains stem entirely from the metacognitive mechanism. The delegation threshold is $\theta = 0.5$, the confidence weight $\lambda = 0.6$, the profile learning rate $\alpha = 0.1$, the conflict threshold $\theta_\delta = 0.3$, and the dampening factor $\gamma = 0.2$.

\textbf{Baselines.}
We compare against six baselines: (1)~\textbf{Single-Agent}: one GPT-4 instance handles all tasks; (2)~\textbf{Round-Robin}: agents take turns processing tasks cyclically; (3)~\textbf{Random-Routing}: tasks randomly assigned to agents; (4)~\textbf{Skill-Fixed}: rule-based routing using keyword matching; (5)~\textbf{Majority-Vote}: all agents solve every task, majority vote determines answer; (6)~\textbf{AutoGen}: the AutoGen framework~\cite{wu2023autogen} with default conversation-based collaboration.

\subsection{Main Results}

Table~\ref{tab:main} presents the main experimental results. MetaCogAgent achieves the highest task accuracy of 82.4\%, surpassing the strongest baseline (Majority-Vote, 77.1\%) by 5.3 percentage points and the best routing baseline (AutoGen, 73.7\%) by 8.7 points. Critically, MetaCogAgent uses only 1382 API calls---5.1\% fewer than AutoGen (1456) while achieving 8.7\% higher accuracy, and 34\% fewer than the brute-force Majority-Vote (2100). While API call counts are a coarse efficiency proxy (actual costs vary with per-call token usage), the reduction demonstrates that intelligent delegation avoids redundant computation.

\begin{table}[t]
\centering
\caption{Main results on MetaCog-Eval. Best in \textbf{bold}.}
\label{tab:main}
\scriptsize
\begin{tabular}{lcccc}
\toprule
\textbf{Method} & \textbf{Acc. (\%)} & \textbf{Del. Prec.} & \textbf{ECE$\downarrow$} & \textbf{API Calls} \\
\midrule
Single-Agent & 65.3 & --- & --- & 700 \\
Round-Robin & 61.8 & --- & --- & 700 \\
Random-Routing & 63.5 & --- & --- & 700 \\
Skill-Fixed & 70.2 & 0.673 & --- & 700 \\
Majority-Vote & 77.1 & --- & --- & 2100 \\
AutoGen & 73.7 & --- & --- & 1456 \\
\midrule
\textbf{MetaCogAgent} & \textbf{82.4} & \textbf{0.841} & \textbf{0.087} & \textbf{1382} \\
\bottomrule
\end{tabular}
\end{table}

A delegation precision of 0.841 means that 84.1\% of delegated tasks are routed to an agent that produces the correct answer. The ECE of 0.087 reflects well-calibrated self-assessment: agents' confidence scores reliably predict actual performance.

\subsection{Difficulty-Stratified Results}

Table~\ref{tab:difficulty} decomposes accuracy by difficulty level. All methods degrade from Easy to Hard, but MetaCogAgent's degradation is the smallest (11.5\% drop vs.\ 18.5\% for Single-Agent). This robustness stems from the adaptive delegation protocol. As tasks become harder and agents' confidence drops, delegation frequency increases (10.5\% $\to$ 41.5\%), routing difficult tasks to the most capable agent rather than forcing execution by the initially assigned one.

\begin{table}[t]
\centering
\caption{Accuracy (\%) by difficulty level.}
\label{tab:difficulty}
\scriptsize
\begin{tabular}{lcccc}
\toprule
\textbf{Method} & \textbf{Easy} & \textbf{Med.} & \textbf{Hard} & \textbf{$\Delta$(E$\to$H)} \\
\midrule
Single-Agent & 76.5 & 63.0 & 58.0 & $-18.5$ \\
Skill-Fixed & 81.0 & 68.5 & 63.0 & $-18.0$ \\
Majority-Vote & 86.0 & 76.5 & 70.0 & $-16.0$ \\
AutoGen & 83.5 & 73.0 & 66.0 & $-17.5$ \\
\textbf{MetaCogAgent} & \textbf{90.5} & \textbf{82.0} & \textbf{79.0} & $\mathbf{-11.5}$ \\
\bottomrule
\end{tabular}
\end{table}

Notably, MetaCogAgent achieves its largest relative improvement on Hard tasks (+13.0\% over AutoGen at Hard vs.\ +7.0\% at Easy), confirming that metacognitive capabilities are most valuable when tasks are challenging---consistent with human behavior, where metacognitive monitoring becomes increasingly important as task difficulty rises~\cite{toppino2009metacognitive}.

\subsection{Performance by Dimension}

\begin{figure}[t]
\centering
\begin{tikzpicture}
\begin{axis}[
    ybar,
    width=\columnwidth,
    height=4.5cm,
    bar width=5pt,
    ylabel={Accuracy (\%)},
    ylabel style={font=\scriptsize},
    symbolic x coords={LR, KR, CG, MC, CI, Cross},
    xtick=data,
    xticklabel style={font=\scriptsize},
    yticklabel style={font=\scriptsize},
    ymin=50, ymax=100,
    legend style={font=\tiny, at={(0.5,1.02)}, anchor=south, legend columns=3},
    nodes near coords style={font=\tiny, rotate=90, anchor=west},
    every node near coord/.append style={xshift=0pt},
    enlarge x limits=0.12,
]
\addplot[fill=gray!40] coordinates {(LR,71.7) (KR,63.3) (CG,62.5) (MC,68.3) (CI,66.7) (Cross,58)};
\addplot[fill=sysblue!40] coordinates {(LR,79.2) (KR,75.8) (CG,72.5) (MC,75.0) (CI,72.5) (Cross,66)};
\addplot[fill=sysgreen!60] coordinates {(LR,85.0) (KR,83.3) (CG,80.8) (MC,84.2) (CI,81.7) (Cross,79)};
\legend{Single-Agent, AutoGen, MetaCogAgent}
\end{axis}
\end{tikzpicture}
\caption{Accuracy by cognitive dimension. MetaCogAgent shows the largest gains on cross-domain tasks requiring multi-skill delegation.}
\label{fig:dimension}
\end{figure}
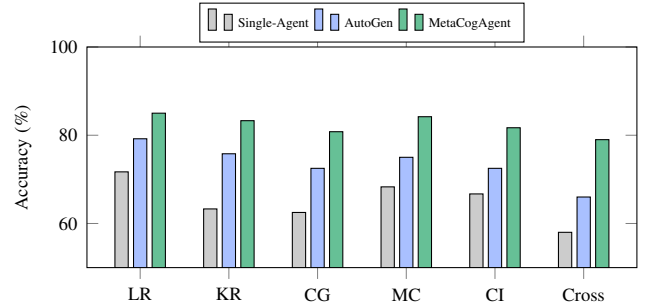

Fig.~\ref{fig:dimension} shows accuracy breakdown across cognitive dimensions. MetaCogAgent's advantage is most pronounced on cross-domain tasks (+13\% over AutoGen), where metacognitive delegation is most critical---these tasks require recognizing that no single agent possesses all necessary skills and routing sub-components to the appropriate specialist.

\subsection{Ablation Study}

Table~\ref{tab:ablation} quantifies the contribution of each metacognitive component. Removing the Self-Assessment module (replacing it with fixed confidence) reduces accuracy by 6.8\%, confirming that dynamic competence estimation is essential. Removing Adaptive Delegation (agents always execute assigned tasks) reduces accuracy by 5.1\%, showing the value of task re-routing. Removing Capability Boundary Learning (static profiles) reduces accuracy by 3.2\%, demonstrating the benefit of cybernetic feedback for profile refinement.

\begin{table}[t]
\centering
\caption{Ablation study on MetaCog-Eval.}
\label{tab:ablation}
\scriptsize
\begin{tabular}{lccc}
\toprule
\textbf{Variant} & \textbf{Acc. (\%)} & \textbf{$\Delta$} & \textbf{Del. Prec.} \\
\midrule
\textbf{MetaCogAgent (Full)} & \textbf{82.4} & --- & \textbf{0.841} \\
\midrule
w/o Self-Assessment & 75.6 & $-6.8$ & 0.612 \\
w/o Adaptive Delegation & 77.3 & $-5.1$ & --- \\
w/o Boundary Learning & 79.2 & $-3.2$ & 0.793 \\
w/o Cross-Agent Eval & 78.9 & $-3.5$ & 0.724 \\
w/o Verbalized Conf. & 78.1 & $-4.3$ & 0.756 \\
\bottomrule
\end{tabular}
\end{table}

\subsection{Confidence Calibration Analysis}

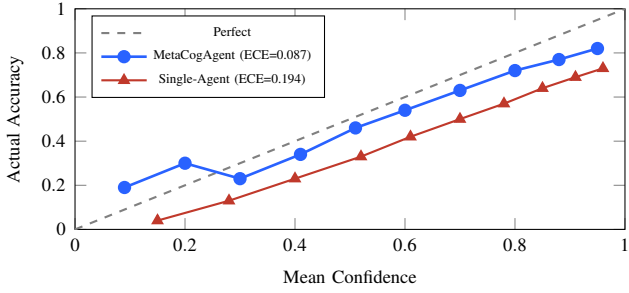
\begin{figure}[t]
\centering
\begin{tikzpicture}
\begin{axis}[
    width=\columnwidth,
    height=4.5cm,
    xlabel={Mean Confidence},
    ylabel={Actual Accuracy},
    xlabel style={font=\scriptsize},
    ylabel style={font=\scriptsize},
    xticklabel style={font=\scriptsize},
    yticklabel style={font=\scriptsize},
    xmin=0, xmax=1, ymin=0, ymax=1,
    legend style={font=\tiny, at={(0.03,0.97)}, anchor=north west},
]
\addplot[dashed, gray, line width=0.8pt] coordinates {(0,0) (1,1)};
\addplot[color=sysblue, mark=*, mark size=2pt, line width=1pt] coordinates {
    (0.09, 0.19) (0.20, 0.30) (0.30, 0.23) (0.41, 0.34)
    (0.51, 0.46) (0.60, 0.54) (0.70, 0.63) (0.80, 0.72)
    (0.88, 0.77) (0.95, 0.82)
};
\addplot[color=sysred, mark=triangle*, mark size=2pt, line width=0.8pt] coordinates {
    (0.15, 0.04) (0.28, 0.13) (0.40, 0.23) (0.52, 0.33)
    (0.61, 0.42) (0.70, 0.50) (0.78, 0.57) (0.85, 0.64)
    (0.91, 0.69) (0.96, 0.73)
};
\legend{Perfect, MetaCogAgent (ECE=0.087), Single-Agent (ECE=0.194)}
\end{axis}
\end{tikzpicture}
\caption{Reliability diagram. MetaCogAgent's confidence is well-calibrated, closely tracking the ideal diagonal. Single-Agent exhibits systematic overconfidence.}
\label{fig:calibration}
\end{figure}

We analyze the calibration quality of MetaCogAgent's self-assessment across difficulty levels. Fig.~\ref{fig:calibration} plots mean confidence against actual accuracy in 10 equal-width bins. The ECE of 0.087 indicates strong calibration overall. Calibration is strongest for Easy tasks (ECE$=$0.051) and degrades for Hard tasks (ECE$=$0.132), suggesting that agents recognize straightforward competence matches more readily than nuanced capability limitations.

\subsection{Delegation Behavior Analysis}

Of the 700 tasks, MetaCogAgent delegated 218 (31.1\%). Delegation rates vary by task difficulty: 10.5\% for Easy, 25.5\% for Medium, and 41.5\% for Hard tasks, demonstrating that the metacognitive system appropriately recognizes increasing task difficulty. Among delegated tasks, 83.5\% were reassigned to agents that achieved higher accuracy than the original assignee, confirming that delegation decisions are well-informed.

Cross-domain tasks exhibit the highest delegation rate (63.0\%), often triggering the Collaborative Mode where multiple agents contribute partial solutions. This emergent behavior---automatically recognizing when a task requires composite expertise---is a direct consequence of metacognitive self-awareness.

Among the 218 delegated tasks, we observe distinct delegation \emph{flow patterns}. Agent-$\beta$ (retrieval-focused) initiates 43.1\% of all delegations, primarily forwarding reasoning-heavy tasks to Agent-$\alpha$. Agent-$\gamma$ (coding) delegates 31.2\% of tasks, mostly mathematical problems to Agent-$\alpha$. Agent-$\alpha$ (reasoning) delegates the least (25.7\%), consistent with reasoning being the most general capability. These asymmetric flow patterns emerge naturally from capability profiles, not from explicit design.

\subsection{Sensitivity Analysis}

We evaluate MetaCogAgent's sensitivity to its three key hyperparameters: delegation threshold $\theta$, confidence weight $\lambda$, and learning rate $\alpha$. Table~\ref{tab:sensitivity} reports accuracy when each parameter is varied while others are held at default values.

\begin{table}[t]
\centering
\caption{Sensitivity analysis. Default values are underlined.}
\label{tab:sensitivity}
\scriptsize
\begin{tabular}{cccccc}
\toprule
\textbf{$\theta$} & \textbf{Acc.} & \textbf{$\lambda$} & \textbf{Acc.} & \textbf{$\alpha$} & \textbf{Acc.} \\
\midrule
0.3 & 78.1 & 0.3 & 79.4 & 0.01 & 80.6 \\
0.4 & 80.9 & 0.5 & 81.8 & 0.05 & 81.7 \\
\underline{0.5} & \textbf{82.4} & \underline{0.6} & \textbf{82.4} & \underline{0.1} & \textbf{82.4} \\
0.6 & 81.3 & 0.7 & 81.6 & 0.2 & 81.9 \\
0.7 & 79.6 & 0.9 & 78.7 & 0.3 & 80.3 \\
\bottomrule
\end{tabular}
\end{table}

The delegation threshold $\theta$ exhibits a clear optimum at 0.5: too low ($\theta{=}0.3$) causes unnecessary delegation of tasks agents could handle competently, wasting API calls; too high ($\theta{=}0.7$) forces agents to attempt tasks beyond their capability. The confidence weight $\lambda$ peaks at 0.6, reflecting the complementary value of combining verbalized confidence with historical profiles. The learning rate $\alpha$ shows the least sensitivity, with performance remaining above 80\% across the tested range, suggesting that the capability boundary learning mechanism is robust to this parameter choice.

\subsection{Qualitative Analysis}

We present a representative example illustrating the metacognitive delegation mechanism. Given a cross-domain task requiring both mathematical reasoning and code generation (``\textit{Write a Python function that computes the expected value of a geometric distribution and verify it against the closed-form solution}''), Agent-$\gamma$ (coding) is initially assigned. Its MCU produces $c_\gamma = 0.38$ ($c^v = 0.45$, $c^p = 0.27$ for math-related tasks), triggering delegation. Cross-evaluation yields $c_\alpha = 0.72$ (Agent-$\alpha$, reasoning) and $c_\beta = 0.21$ (Agent-$\beta$, retrieval). Agent-$\alpha$ executes the task, producing a correct solution with both the derivation and implementation. Without metacognition, Agent-$\gamma$ would have attempted the task directly, producing syntactically valid but mathematically incorrect code---a subtle failure mode that downstream systems would struggle to detect.

\section{Discussion}

\textbf{When does metacognition help most?}
Our results reveal a clear pattern: metacognitive capabilities provide the largest benefits when (a) tasks are difficult (Hard: +13\% over AutoGen vs.\ Easy: +7\%), (b) tasks require cross-domain expertise (+13\% on cross-domain), and (c) the initially assigned agent is poorly matched to the task. For easy, well-matched tasks, MetaCogAgent's overhead (confidence evaluation) provides marginal benefit over simpler routing. This suggests that practical deployments could use a lightweight pre-filter to bypass metacognitive evaluation for routine tasks, reducing computational costs further.

\textbf{Emergent specialization.}
An unexpected finding is that capability profiles converge to reflect genuine agent specialization patterns. After processing all 700 tasks, Agent-$\alpha$'s profile shows high competence on LR (0.89) and MC (0.85) but lower on CG (0.62), while Agent-$\gamma$ shows the inverse pattern (CG: 0.87, LR: 0.64). These profiles emerge purely from feedback---no explicit specialization was encoded beyond the initial system prompts. This suggests that cybernetic feedback can drive \emph{functional specialization} in multi-agent systems, a phenomenon relevant to the SMC community's interest in self-organizing systems.

\textbf{Relationship to human metacognition.}
Our framework models two of the three metacognitive components identified by Flavell~\cite{flavell1979metacognition}: metacognitive knowledge (capability profiles) and metacognitive monitoring (self-assessment). The third component---metacognitive control (strategic planning of cognitive resources)---is partially addressed by the delegation protocol but could be extended to include more sophisticated strategies such as task decomposition and selective attention.

\textbf{Generalizability beyond MetaCog-Eval.}
Our benchmark is purpose-built to isolate metacognitive behaviors (self-assessment accuracy, delegation quality) that standard benchmarks like MMLU or HumanEval do not measure. Nevertheless, the underlying mechanism---confidence-gated delegation---is domain-independent in principle, though its effectiveness depends on the degree of competence heterogeneity among agents. We expect the largest gains under strong capability asymmetry and the smallest when agents are near-identical.

\section{Conclusion}

We presented MetaCogAgent, a multi-agent LLM framework that introduces metacognitive capabilities inspired by human self-awareness. By enabling agents to assess their own competence boundaries and adaptively delegate tasks they cannot handle, MetaCogAgent achieves 82.4\% accuracy on our MetaCog-Eval benchmark, outperforming the best routing baseline by 8.7\% while using 5\% fewer API calls than AutoGen and 34\% fewer than ensemble voting. The cybernetic feedback loop for capability boundary learning ensures that the system progressively improves its delegation decisions, with sensitivity analysis confirming robustness across hyperparameter settings.

\textbf{Limitations.} The current framework evaluates metacognition with three agents; scalability to larger agent populations warrants further investigation. The capability profile learning assumes stationary competence distributions; handling non-stationary environments where agent capabilities change (e.g., due to prompt modifications or model updates) remains an open challenge. Additionally, our benchmark relies on GPT-4 for both task generation and agent execution, which may introduce distributional biases; evaluation with heterogeneous backbone models (e.g., mixing GPT-4 with open-source LLMs) on public benchmarks would further validate generalizability. All reported results are single-run; future work should include variance estimates.

\textbf{Future Work.} We plan to extend MetaCogAgent to open-ended domains where agents discover their own competence categories, and to explore hierarchical metacognition where a meta-agent monitors individual agents' self-assessment quality~\cite{toppino2009metacognitive}. The MetaCog-Eval benchmark and code will be released upon publication.

\balance
\bibliographystyle{IEEEtran}
\bibliography{references}

@inproceedings{wu2023autogen,
  title={Auto{G}en: Enabling Next-Gen {LLM} Applications via Multi-Agent Conversation},
  author={Wu, Qingyun and Bansal, Gagan and Zhang, Jieyu and Wu, Yiran and Li, Beibin and Zhu, Erkang and Jiang, Li and Zhang, Xu and Wang, Shaokun and Zhang, Sheng and others},
  booktitle={Proceedings of the International Conference on Machine Learning (ICML)},
  year={2024}
}

@article{hong2023metagpt,
  title={Meta{GPT}: Meta Programming for A Multi-Agent Collaborative Framework},
  author={Hong, Sirui and Zhuge, Mingchen and Chen, Jonathan and Zheng, Xiawu and Cheng, Yuheng and Zhang, Ceyao and Wang, Jinlin and Wang, Zili and Yau, Steven Ka Shing and Lin, Zijuan and others},
  journal={arXiv preprint arXiv:2308.00352},
  year={2023}
}

@article{flavell1979metacognition,
  title={Metacognition and Cognitive Monitoring: A New Area of Cognitive-Developmental Inquiry},
  author={Flavell, John H},
  journal={American Psychologist},
  volume={34},
  number={10},
  pages={906--911},
  year={1979}
}

@article{kadavath2022language,
  title={Language Models (Mostly) Know What They Know},
  author={Kadavath, Saurav and Conerly, Tom and Askell, Amanda and Henighan, Tom and Drain, Dawn and Perez, Ethan and Schiefer, Nicholas and Hatfield-Dodds, Zac and DasSarma, Nova and Tran-Johnson, Eli and others},
  journal={arXiv preprint arXiv:2207.05221},
  year={2022}
}

@inproceedings{yin2023exchange,
  title={Exchange-of-Thought: Enhancing Large Language Model Capabilities through Cross-Model Communication},
  author={Yin, Zhangyue and Sun, Qiushi and Chang, Cheng and Guo, Qipeng and Dai, Junqi and Huang, Xuanjing and Qiu, Xipeng},
  booktitle={Proceedings of the Conference on Empirical Methods in Natural Language Processing (EMNLP)},
  year={2023}
}

@inproceedings{du2023improving,
  title={Improving Factuality and Reasoning in Language Models through Multiagent Debate},
  author={Du, Yilun and Li, Shuang and Torralba, Antonio and Tenenbaum, Joshua B and Mordatch, Igor},
  booktitle={Proceedings of the International Conference on Machine Learning (ICML)},
  year={2024}
}

@article{xiong2024llms,
  title={Can {LLMs} Express Their Uncertainty? An Empirical Evaluation of Confidence Elicitation in {LLMs}},
  author={Xiong, Miao and Hu, Zhiyuan and Lu, Xinyang and Li, Yifei and Fu, Jie and He, Junxian and Hooi, Bryan},
  journal={arXiv preprint arXiv:2306.13063},
  year={2024}
}

@inproceedings{guo2017calibration,
  title={On Calibration of Modern Neural Networks},
  author={Guo, Chuan and Pleiss, Geoff and Sun, Yu and Weinberger, Kilian Q},
  booktitle={Proceedings of the International Conference on Machine Learning (ICML)},
  pages={1321--1330},
  year={2017}
}

@inproceedings{li2023camel,
  title={{CAMEL}: Communicative Agents for ``Mind'' Exploration of Large Language Model Society},
  author={Li, Guohao and Hammoud, Hasan Abed Al Kader and Itani, Hani and Khizbullin, Dmitrii and Ghanem, Bernard},
  booktitle={Advances in Neural Information Processing Systems (NeurIPS)},
  year={2023}
}

@book{wiener1948cybernetics,
  title={Cybernetics: Or Control and Communication in the Animal and the Machine},
  author={Wiener, Norbert},
  publisher={MIT Press},
  address={Cambridge, MA},
  year={1948}
}

@article{toppino2009metacognitive,
  title={Metacognitive Control and Strategy Selection: Deciding to Practice Retrieval During Learning},
  author={Toppino, Thomas C and Cohen, Michael S},
  journal={Journal of Experimental Psychology: Learning, Memory, and Cognition},
  volume={35},
  number={5},
  pages={1105--1117},
  year={2009}
}

@inproceedings{park2023generative,
  title={Generative Agents: Interactive Simulacra of Human Behavior},
  author={Park, Joon Sung and O'Brien, Joseph C and Cai, Carrie J and Morris, Meredith Ringel and Liang, Percy and Bernstein, Michael S},
  booktitle={Proceedings of the ACM Symposium on User Interface Software and Technology (UIST)},
  year={2023}
}

@inproceedings{chen2024agentverse,
  title={Agent{V}erse: Facilitating Multi-Agent Collaboration and Exploring Emergent Behaviors},
  author={Chen, Weize and Su, Yusheng and Zuo, Jingwei and Yang, Cheng and Yuan, Chenfei and Chan, Chi-Min and Yu, Heyang and Lu, Yaxi and Hung, Yi-Hsin and Qian, Chen and others},
  booktitle={International Conference on Learning Representations (ICLR)},
  year={2024}
}

@inproceedings{brown2020language,
  title={Language Models are Few-Shot Learners},
  author={Brown, Tom and Mann, Benjamin and Ryder, Nick and Subbiah, Melanie and Kaplan, Jared D and Dhariwal, Prafulla and Neelakantan, Arvind and Shyam, Pranav and Sastry, Girish and Askell, Amanda and others},
  booktitle={Advances in Neural Information Processing Systems (NeurIPS)},
  volume={33},
  pages={1877--1901},
  year={2020}
}

@article{achiam2023gpt4,
  title={{GPT-4} Technical Report},
  author={Achiam, Josh and Adler, Steven and Agarwal, Sandhini and Ahmad, Lama and Akkaya, Ilge and Aleman, Florencia Leoni and Almeida, Diogo and Altenschmidt, Janko and Altman, Sam and Anadkat, Shyamal and others},
  journal={arXiv preprint arXiv:2303.08774},
  year={2023}
}

@inproceedings{shinn2023reflexion,
  title={Reflexion: Language Agents with Verbal Reinforcement Learning},
  author={Shinn, Noah and Cassano, Federico and Gopinath, Ashwin and Narasimhan, Karthik and Yao, Shunyu},
  booktitle={Advances in Neural Information Processing Systems (NeurIPS)},
  year={2023}
}

@inproceedings{yao2023tree,
  title={Tree of Thoughts: Deliberate Problem Solving with Large Language Models},
  author={Yao, Shunyu and Yu, Dian and Zhao, Jeffrey and Shafran, Izhak and Griffiths, Thomas L and Cao, Yuan and Narasimhan, Karthik},
  booktitle={Advances in Neural Information Processing Systems (NeurIPS)},
  year={2023}
}

\end{document}